\documentclass[runningheads]{llncs}
\usepackage{comment}
\usepackage{graphicx}
\usepackage{amsmath}
\usepackage{cite}
\usepackage{booktabs}

\begin{document}
\title{Confronting the Constraints for Optical Character Segmentation from Printed Bangla Text Image}
\titlerunning{Confronting the Constraints of Bangla OCR System}
%
\author{Abu Saleh Md. Abir \and
Sanjana Rahman \and
Samia Ellin \and
Maisha Farzana \and
Md. Hridoy Manik \and
Chowdhury Rafeed Rahman
}
\authorrunning{Abir et al.}
%
\institute{United International University, Dhaka, Bangladesh}
\maketitle              
\begin{abstract}
In a world of digitization, optical character recognition holds the automation to written history. Optical character recognition system basically converts printed images into editable texts for better storage and usability. To be completely functional, the system needs to go through some crucial methods such as pre-processing and segmentation. Pre-processing helps printed data to be noise free and gets rid of skewness efficiently whereas segmentation helps the image fragment into line, word and character precisely for better conversion. These steps hold the door to better accuracy and consistent results for a printed image to be ready for conversion. Our proposed algorithm is able to segment characters both from ideal and non-ideal cases of scanned or captured images giving a sustainable outcome. The implementation of our work is provided here: \textbf{https://cutt.ly/rgdfBIa}.

\end{abstract}
\section{Introduction}
\begin{figure}[ht]
  \centering
  \includegraphics[width=100mm]{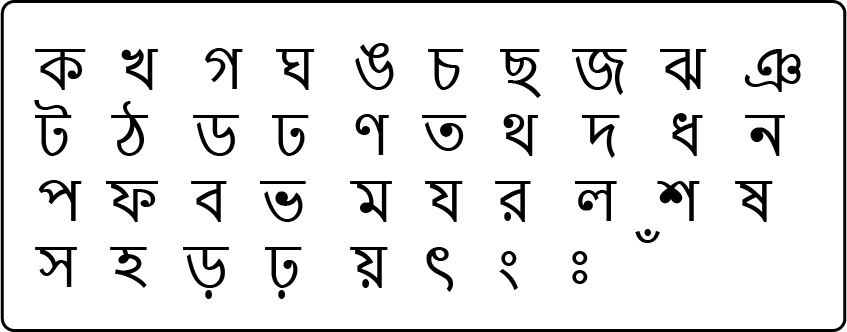}
  \caption{Bangla Consonants.}
    \label{fig: Consonant}
\end{figure}

\begin{figure}[ht]
  \centering
  \includegraphics[width=100mm]{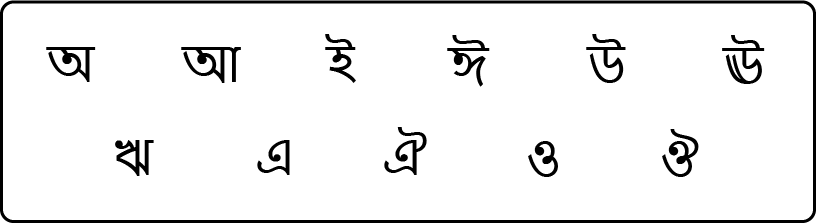}
  \caption{Bangla Vowels.}
    \label{fig: Vowel}
\end{figure}

\begin{figure}[ht]
  \centering
  \includegraphics[width=100mm]{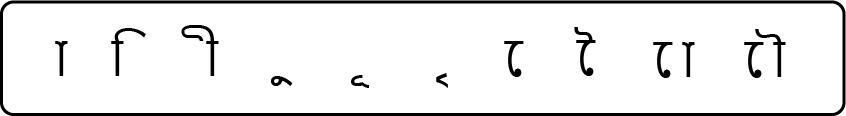}
  \caption{Bangla Modified Vowels.}
    \label{fig: Modifier}
\end{figure}

\begin{figure}[ht]
  \centering
  \includegraphics[width=75mm]{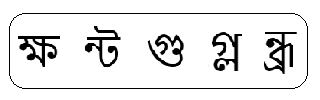}
  \caption{Some Bangla Compound Characters.}
    \label{fig: Compounds}
\end{figure}

Optical character recognition (OCR) is the process of converting printed text images into editable texts. As physical media such as books, newspapers, important files can get destroyed easily or can be damaged after a certain period of time, converting them into a more persistent media is the only option. OCR system works as a digital media that can store valuable information from the physical media in an effective way. It is the key towards a better mechanism which can be time-efficient, effortless and productive. Though Bangla is a popular language, it does not have a proper OCR system compared to other languages such as English. Bangla as a language is complex and the writing structure is different from other languages. Bangla language has consonants (Fig. \ref{fig: Consonant}), vowels (Fig. \ref{fig: Vowel}), modified vowels (Fig. \ref{fig: Modifier}) and around 170 compound characters (Fig. \ref{fig: Compounds}) \cite{1}. Such complex writing structure needs better segmentation process for conversion into digital media, hence the applications for it is difficult.
 
For an OCR system to work properly, we need to segment each of the characters properly. A printed text image needs to be pre-processed and segmented properly before it can be converted into editable text. The main challenge is to prepare the image for segmentation. It is the pre-processing phase that gets rid of skewness of image, straighten the curved lines, eliminate unwanted noise in the image and many more. For segmentation to work, these obstacles need to be removed with care. Then for an image to be prepared for transforming into editable text, the text needs to be segmented into lines, words and characters accurately.

Given an image as input, we provide the segmented images of each character of the input image along with each segmented character's line number and word number in that page as output. Our method works well even if the image capture condition is far from ideal. We do not tackle the challenges of character recognition in this research. When an image is scanned, it enters the pre-processing stage. At first the image is cropped to remove any borders around the text. Then the orientation of the image is corrected and an algorithm is used to correct the warped images and straighten the curved text lines of the image. If there is any noise in the image, it is removed. Finally, binarization is performed to convert the pixel values of the image into 1s and 0s. After the pre-processing is done, the image goes through segmentation phase. At first the image is scanned horizontally and each of the lines are segmented. Then each of the line is scanned vertically to segment each of the words. Finally, the segmented words are used to segment each of the characters to give the final outcome.

We have reviewed Bangla OCR researches from the perspective of limitation finding and have implemented them to gain full understanding of the challenges associated with Bangla OCR development. We have developed our own approach to overcome the current limitations of pre-processing and segmentation phase of Bangla OCR system. The accuracy of character recognition and sentence reconstruction depend largely on the precision of these phases. We have collected images of many non-ideal cases and have confronted the common obstacles in this research. Results of our algorithm on such non-ideal case images show the effectiveness of our method.

\section{Literature Review}
Some problem scopes of Bangla OCR system are mentioned in \cite{2}, which include lack of standard samples and complex structure of documents. One of the drawbacks of the OCR system is that it does not work properly if the resolution of the image is less than 300dpi \cite{3, 4}. A printed Bangla OCR system was developed using a single hidden BLSTM-CTC architecture which includes pre-processing, line detection and recognition \cite{4}. But the system only works for fixed font size which was used to train the model. One of the approaches include two zone approach for character segmentation \cite{5}, which do not work for all fonts and sizes. The accuracy of the model was reduced due to some connected characters. A group of researchers suggested a complete Bangla OCR system methodology where they experimented their methods for different fonts and sizes and got a good accuracy rate only for larger font size \cite{6}. Tesseract is an open source OCR engine which works for many languages \cite{7}. Recently, it has started to work on Bangla script as well. But one drawback of this OCR engine is that it does not work for all Bangla fonts and sizes. An effective way of a complete model for Bangla OCR system was proposed in \cite{8}, but it was performed only for three fixed fonts and no compound characters were included in training data. A research showed some degradation of printed Bangla script due to connected characters, broken characters and for light or heavy printed documents \cite{9}.

A proposed method for character segmentation in a handwritten document suggests vertex characterization of outer isothetic polygonal covers. This method shows poor performance for some pathological cases, as it suffers from under-segmentation and over-segmentation problem \cite{10}. For improving the character segmentation accuracy, a new algorithm was proposed using back propagation neural network \cite{11}, though the accuracy was not good. A high performance OCR for Bangla script was presented in \cite{12} which works only for some specific documents and articles. But the proposed method has problems with segmentation which lowers the accuracy of the system. Another OCR system was proposed to recognize the segmented characters by using an artificial neural network in \cite{13}. But this approach shows poor accuracy for character segmentation and fails to recognize similar Bangla characters. 

In order to improve the output result of Bangla OCR system, different correction algorithms were proposed. The main problem of these researches is the misclassification of some compound characters \cite{14}. To recognize these compound characters, a deep CNN with RELU as nonlinear function was proposed in \cite{15}, but it only works for fixed data-set and cannot recognize all the character classes. Performance evaluation of different algorithms related to Bangla character recognition has been presented in \cite{16}, though this approach misclassifies some similar characters. Another recognition method \cite{17} was proposed for handwritten Bangla numeral which includes three variants of Local Binary Pattern (LBP) using neural network. They compared the performances of these three variants.

\section{Ideal Case Vs. Non-Ideal Case Scenario}
\begin{figure}[ht]
  \centering
  \includegraphics[width=\linewidth]{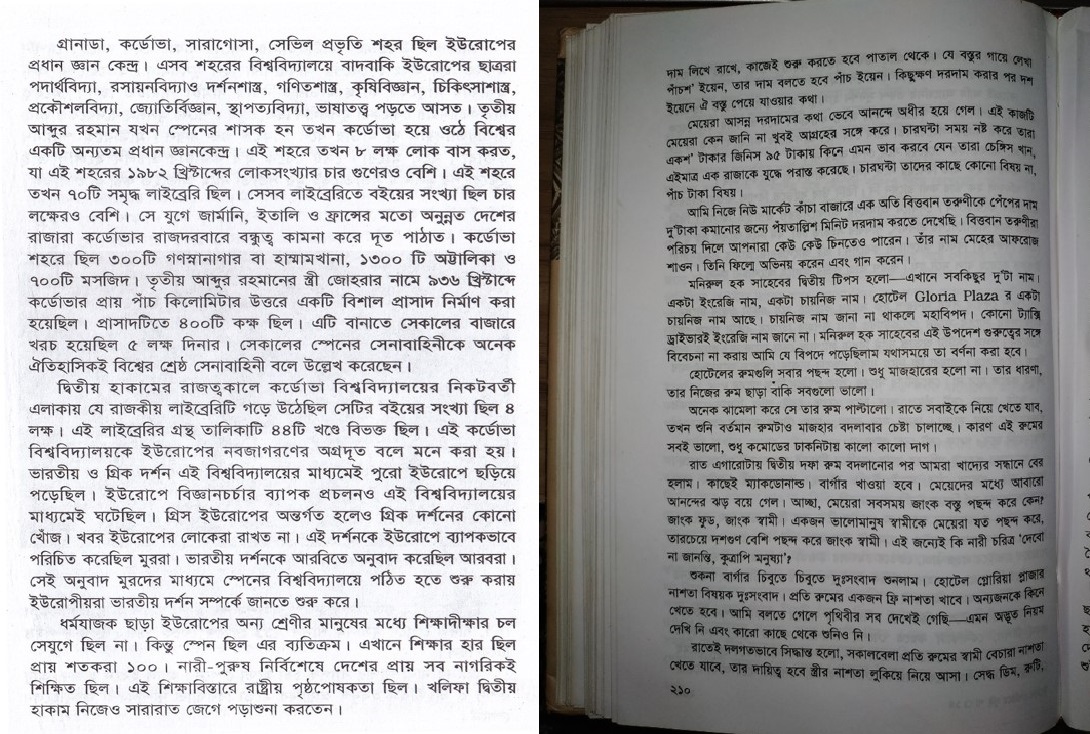}
  \caption{Ideal Case (left) and Non-ideal Case (right) Scenario.}
    \label{fig: BestWorse}
\end{figure}

Many of the researches performed on Bangla OCR system have worked with ideal case scenarios only. Fig. \ref{fig: BestWorse} shows example of an ideal and a non-ideal case scenario. An ideal case image is cropped precisely from the border of the page and the text lines are straight. The image is not skewed and there is also no noise in the image. A non-ideal image may have portions on the image outside of the page and needs to be cropped. The text lines are not always straight and the page may be warped and may contain noises. So it is easier to work with ideal case images and segment the lines, words, and characters from it. Our proposed algorithm performance is satisfactory even for the non-ideal cases.

\section{Methodology}

\begin{figure}[ht]
  \centering
  \includegraphics[width=\linewidth]{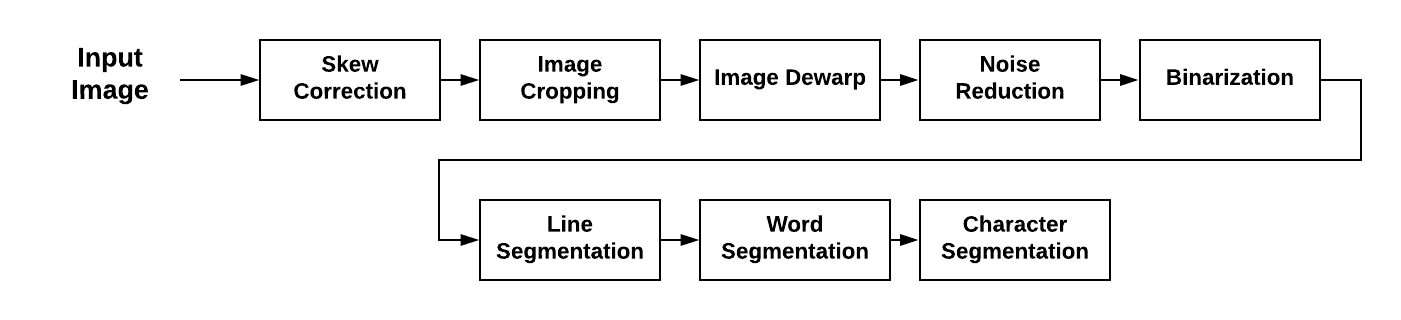}
  \caption{OCR Pre-Processing and Segmentation Phases.}
    \label{fig: Methods}
\end{figure}

We tackle the limitations of the steps shown in Fig. \ref{fig: Methods} in this research through smart algorithm design. Each step of our algorithm has been described as follows:

\subsection{Skew Correction}
\begin{figure}[ht]
  \centering
  \includegraphics[width=100mm]{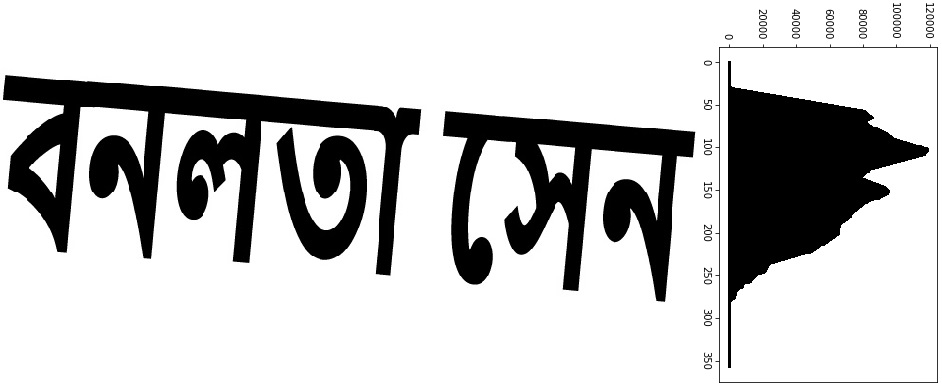}
  \caption{Skewed Line Alongside Row-wise Pixel Sum Graph.}
    \label{fig: angle}
\end{figure}

Input images may be skewed due to position of the camera or due to the position of the book on the scanner. Many of the Bangla OCR systems fail in this common scenario. It is important to rotate the image to an angle where the image is vertically straight. 

\begin{figure}[ht]
  \centering
  \includegraphics[width=100mm]{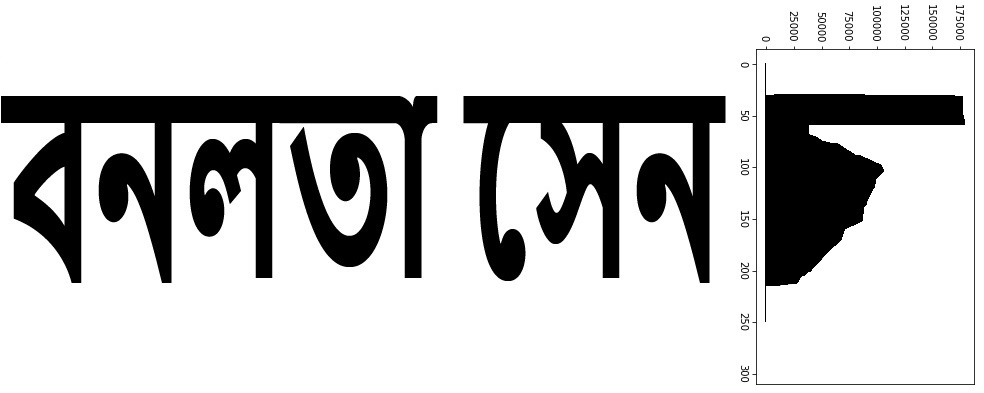}
  \caption{Corrected Skewed Line alongside Row-wise Pixel Sum Graph.}
    \label{fig: straight}
\end{figure}

\begin{figure}[ht]
  \centering
  \includegraphics[width=120mm]{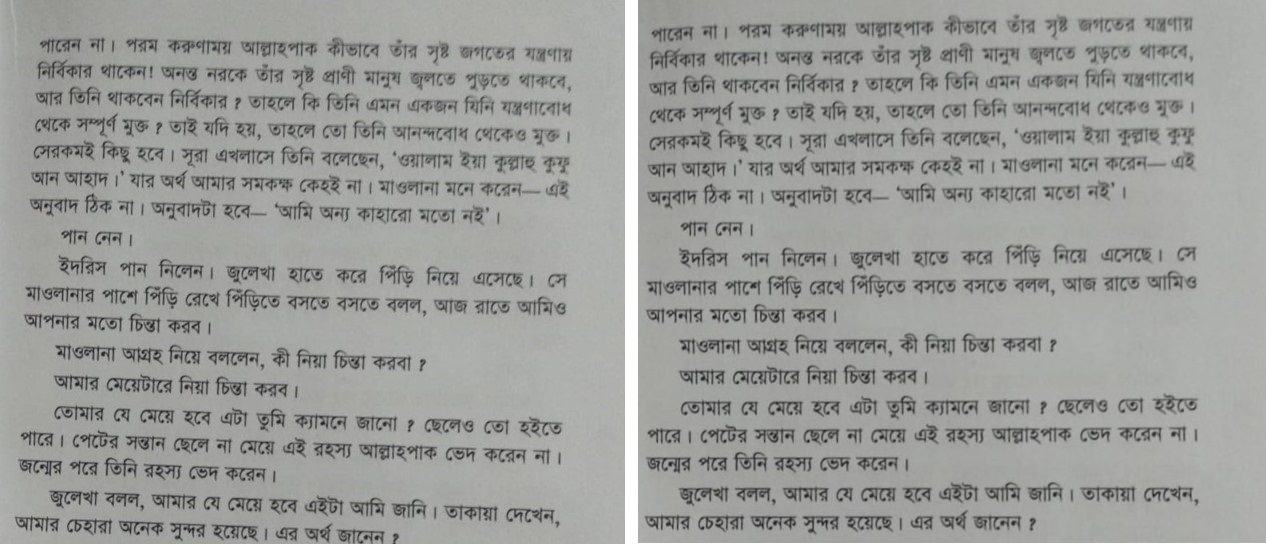}
  \caption{Before (left) and After (right) Image Skew Correction Execution.}
    \label{fig: Skew}
\end{figure}

We have calculated the sum of each row of pixels to generate a row-wise pixel sum histogram. Then the image is inter rotated in each iteration from a range of -5 to 5 degree angle. To determine the skew angle, we compared the maximum difference between the peaks of the rotated angle of images. The skewness of the image is corrected with the angle where the row-wise pixel sum is maximum. As correcting the skewness of the image will provide the matra line row with maximum pixel sum. Fig. \ref{fig: angle} shows an image with a skewed line alongside its row-wise pixel sum graph where the maximum row-wise pixel sum value of the text is 120000. Fig. \ref{fig: straight} shows a corrected skewed line text alongside its row-wise pixel sum graph where the maximum row-wise pixel sum value is 175000. With the maximum row-wise sum pixel of the rotated image, the best angle for the image skew correction is found. Fig. \ref{fig: Skew} shows an example of an image before and after skew correction.

\subsection{Image Cropping}
\begin{figure}[ht]
  \centering
  \includegraphics[width=100mm]{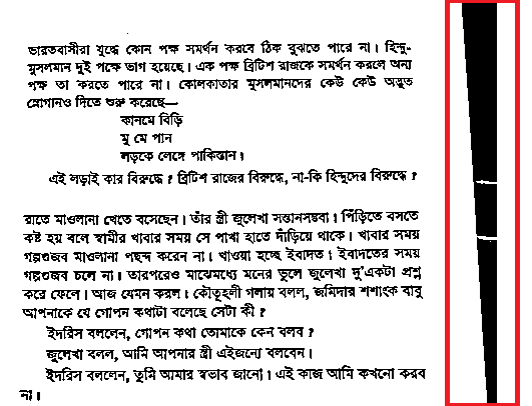}
  \caption{Unwanted Chunk of Black Pixels.}
    \label{fig: UnwantedChunk}
\end{figure}

When an image is captured from a book or document, the image may contain some portion outside of the text page. One of the challenges here is to identify the text of the image and crop the image so that the unwanted parts outside of the text can be eliminated. When binarized, these unwanted parts provide a chunk of black pixels which result in poor segmentation of lines. Fig. \ref{fig: UnwantedChunk} shows unwanted chunk of black pixels marked in a red box.

\begin{figure}
  \centering
  \includegraphics[width=140mm]{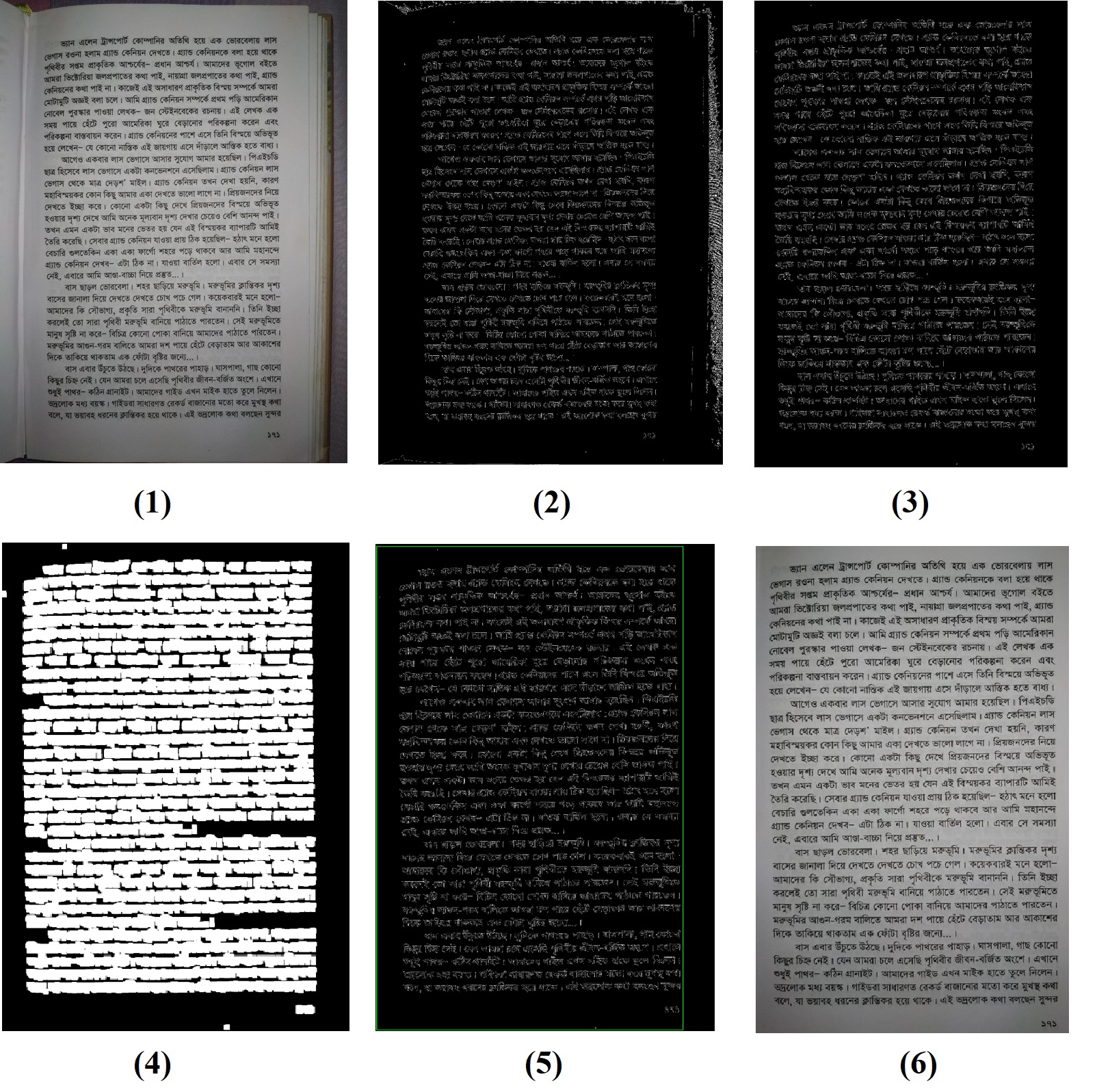}
  \caption{Steps of Text Image Significant Portion Cropping.}
    \label{fig: Cropped}
\end{figure}

To solve the problem, the input image is cropped. For appropriate cropping of the image, we have performed page-layout analysis and have found out where the text is. Fig. \ref{fig: Cropped} shows step by step procedures of how a text image is cropped which have been described below: 

\begin{enumerate}
  \item We are given a casually captured input image containing text.
  \item We use canny edge detection to detect all the pixels wherever, there is an edge. The source of edges in the image are the borders of the page and the text.
  \item To remove the borders, we use rank filters. The text areas have lots of white pixels, but the borders consist of thin single pixel lines. Rank filter replaces these pixels with the median of the pixels. This eliminates the one pixel lines or edges after applying a vertical and horizontal rank filter removing the border pixels.
  \item The next step is to find the contours of the connected components which will be only of the text.
  \item We form a bounding box around the text and crop the image.
  \item Our output is the heterogeneous background free cropped image containing only text. 
\end{enumerate}

\subsection{Image Dewarping}
\begin{figure}[ht]
  \centering
  \includegraphics[width=120mm]{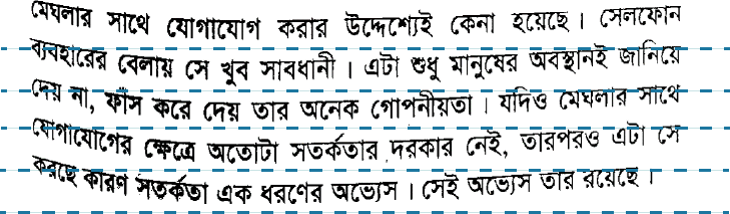}
  \caption{Curved Lines Due to View Angle of Camera or Warped Page.}
    \label{fig: CurvedLines}
\end{figure}

Image dewarping is associated to perspective correction. Geometric distortion of a captured image lines is a common real life scenario. The formation of curved lines due to view angle of camera or warped page leads to poor line segmentation, as most of the lines overlap with each other. As a result, multiple lines get segmented as a single line. Fig. \ref{fig: CurvedLines} shows such a problematic scenario.
  
The steps of our proposed image dewarp algorithm are as follows.
\begin{itemize}
  \item We obtain page boundaries, which consist of the four corners of the image that we have cropped earlier.
  \item	We detect text contours using connected component analysis. Our procedure starts with text dilation which results in connection of neighbouring words situated in the same line. Then each of the lines is detected as connected component and is assembled as a span. 
  \item	Each span of text is then remapped with a calculated parameter estimation. The parameter estimation is calculated depending on the shape of the span and an angle that it is distorted at. With the estimated parameter, coordinate transformation is done to make the lines parallel and horizontal.
  \item Finally, we optimize the remapping of span to minimize the re-projection error using scipy.optimize.minimize which is a derivative-free optimizer.
\end{itemize}

\begin{figure}[ht]
  \centering
  \includegraphics[width=130mm]{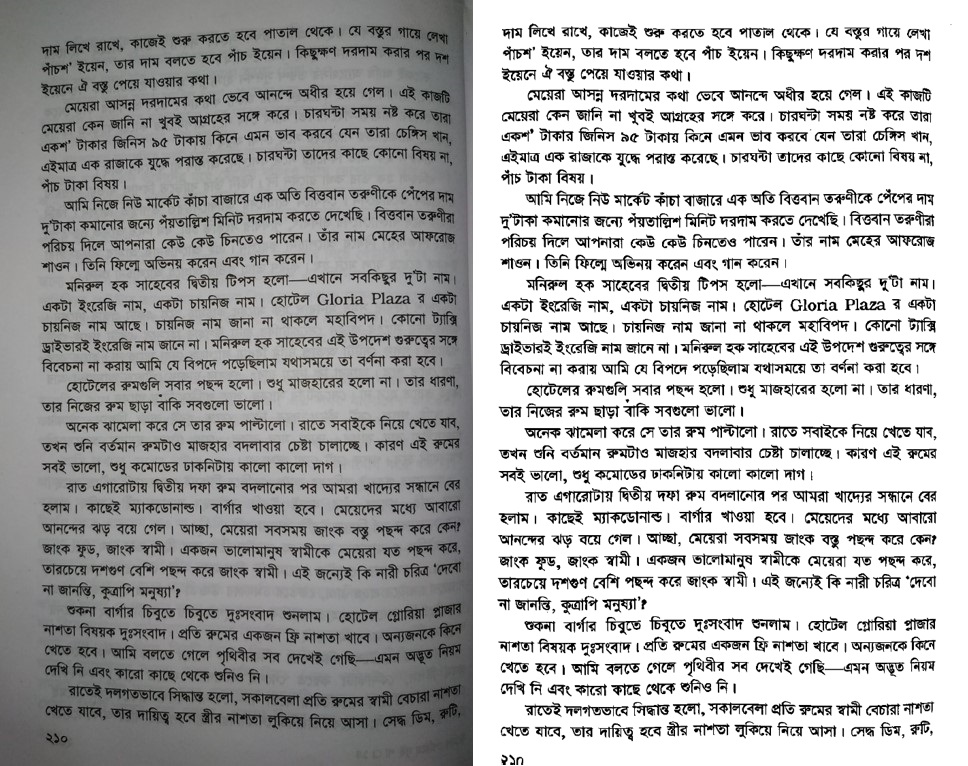}
  \caption{Before (left) and After (right) Image Dewarp Execution.}
    \label{fig: Dewarp}
\end{figure}

Fig. \ref{fig: Dewarp} shows an example of a geometrically distorted image before and after the use of image dewarping.

\subsection{Noise Reduction}
Images can have unwanted noise in them. Noises are mainly of two types. \textbf{Salt and pepper noise} is created due to sudden disturbances in the image, while \textbf{background noise} occurs due to poor intensity during image capture. Since the Bangla characters are complex, this noise reduction step is essential. Otherwise, some of the noises can be assumed as part of the characters. We use a Gaussian smoothing filter based denoising technique to eliminate these noises.

\subsection{Binarization}
Binarization is performed to convert the pixel values of the image to either 1 or 0. This step is very important as it helps us to distinguish between the text and the background when we perform segmentation. First the image is converted into a gray-scale image and an adaptive threshold algorithm is used for binarization of the image.

\subsection{Line Segmentation}
\begin{figure}[ht]
  \centering
  \includegraphics[width=100mm]{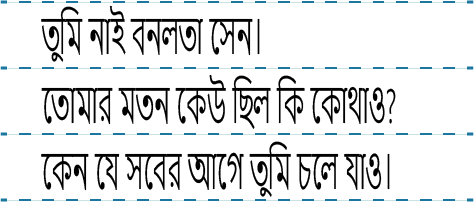}
  \caption{Line Segmentation.}
    \label{fig: LineSegment}
\end{figure}

Lines are detected easily from an image horizontally. At first the image pixel values are calculated for each of the rows and are compared. Line segmentation is performed where the sum of the pixel value is close to zero (Fig. \ref{fig: LineSegment}).

\begin{figure}[ht]
  \centering
  \includegraphics[width=100mm]{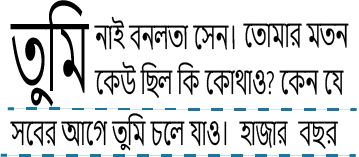}
  \caption{Problem with Multiple Font Size in Line Segmentation.}
    \label{fig: MultipleFont}
\end{figure}

Here, we have faced another challenge while working with multiple font sizes in single page, where we fail to segment each of the lines properly. When multiple font sizes are present on the same image, line segmentation is performed for the bigger font size. As a result, all the lines are not segmented correctly. As shown in Fig. \ref{fig: MultipleFont}, the first two lines get segmented together due to different font sizes.

We resolve this problem using a smart line segmentation technique. The steps of this technique are as follows. 
\begin{itemize}
\item At first, we segment all the lines and check if all the lines are of the same height or not.
\item If any segmented line has height greater than the average height, that segmented line image contains multiple size fonts. To separate the different fonts, we scan and segment the image Vertically which will separate the different fonts.
\item Line segmentation is performed again on the two segmented images to separate any lines that were segmented together due to multiple fonts. 
\item The larger font text portion is resized and is attached with the first segmented line.
\end{itemize}

\subsection{Word Segmentation}
\begin{figure}[ht]
  \centering
  \includegraphics[width=100mm]{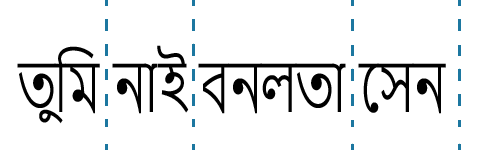}
  \caption{Word Segmentation.}
    \label{fig: WordSegment}
\end{figure}

  Words are segmented easily from segmented line images in a vertical manner. At first, the image pixel values are calculated for each of the columns of a segmented line image. Word segmentation is performed where the sum of the pixel value is close to zero (Fig. \ref{fig: WordSegment}). 

\subsection{Character Segmentation}
Character segmentation is performed on segmented word images. It is the most difficult part among the three segmentation phases. Bangla characters are connected with each other with a headline known as \textbf{matra line}. Bangla characters also have some overlapping modified vowels with them which make this segmentation process more complicated. Separating these overlapping characters and connecting them back again make the task difficult. Character segmentation is divided into two parts - removal of the detected matra line and segmentation of each character. 

\begin{figure}[ht]
  \centering
  \includegraphics[width=80mm]{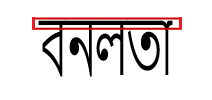}
  \caption{Detection of Matra Line Region.}
    \label{fig: MatraDetection}
\end{figure}

To separate each character, we need to detect matra line and then remove it. To properly detect the matra line, we have horizontally divided the word image into half. Matra line is detected where the sum of pixel value of rows are greater than 60\% on the upper half of the image. Fig. \ref{fig: MatraDetection} shows the region of Matra line for a word.

\begin{figure}[ht]
  \centering
  \includegraphics[width=55mm]{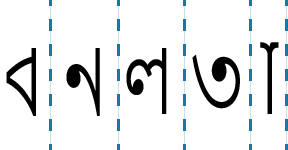}
  \caption{Character Segmentation.}
    \label{fig: CharSegment}
\end{figure}

\begin{figure}[ht]
  \centering
  \includegraphics[width=80mm]{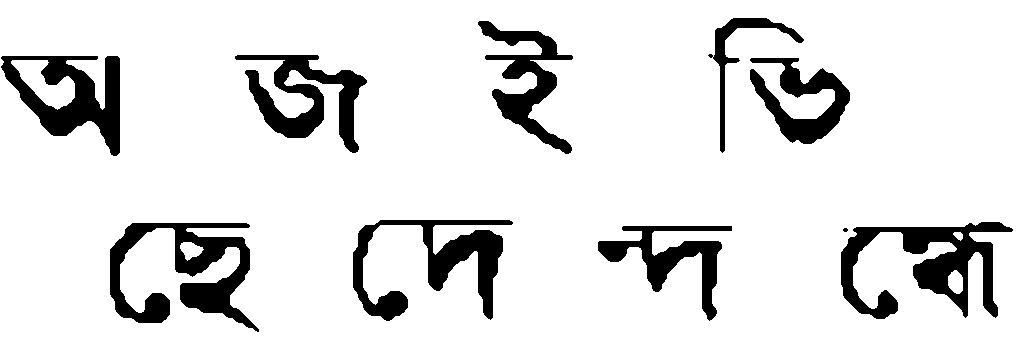}
  \caption{Correctly Segmented Characters.}
    \label{fig: CorrectChar}
\end{figure}

After removing the matra line, we get an open space between the characters as the characters are not connected with each other with the matra line. Characters then can be detected from an image vertically. At first the image pixel values are calculated for each of the columns. Character segmentation is performed where the sum of the pixel values is close to zero (Fig. \ref{fig: CharSegment}). This is because, in a few cases, matra line is removed partially. Sometimes only a portion of matra line gets removed for which segmentation is done with sum of pixel value close to zero. Character segmentation is considered to be correct if a consonant or a vowel or a compound character is segmented alone or alongside with a modified vowel. Fig. \ref{fig: CorrectChar} shows examples of some correctly segmented characters.

\section{Results and Discussion}
\begin{figure}[ht]
  \centering
  \includegraphics[width=140mm]{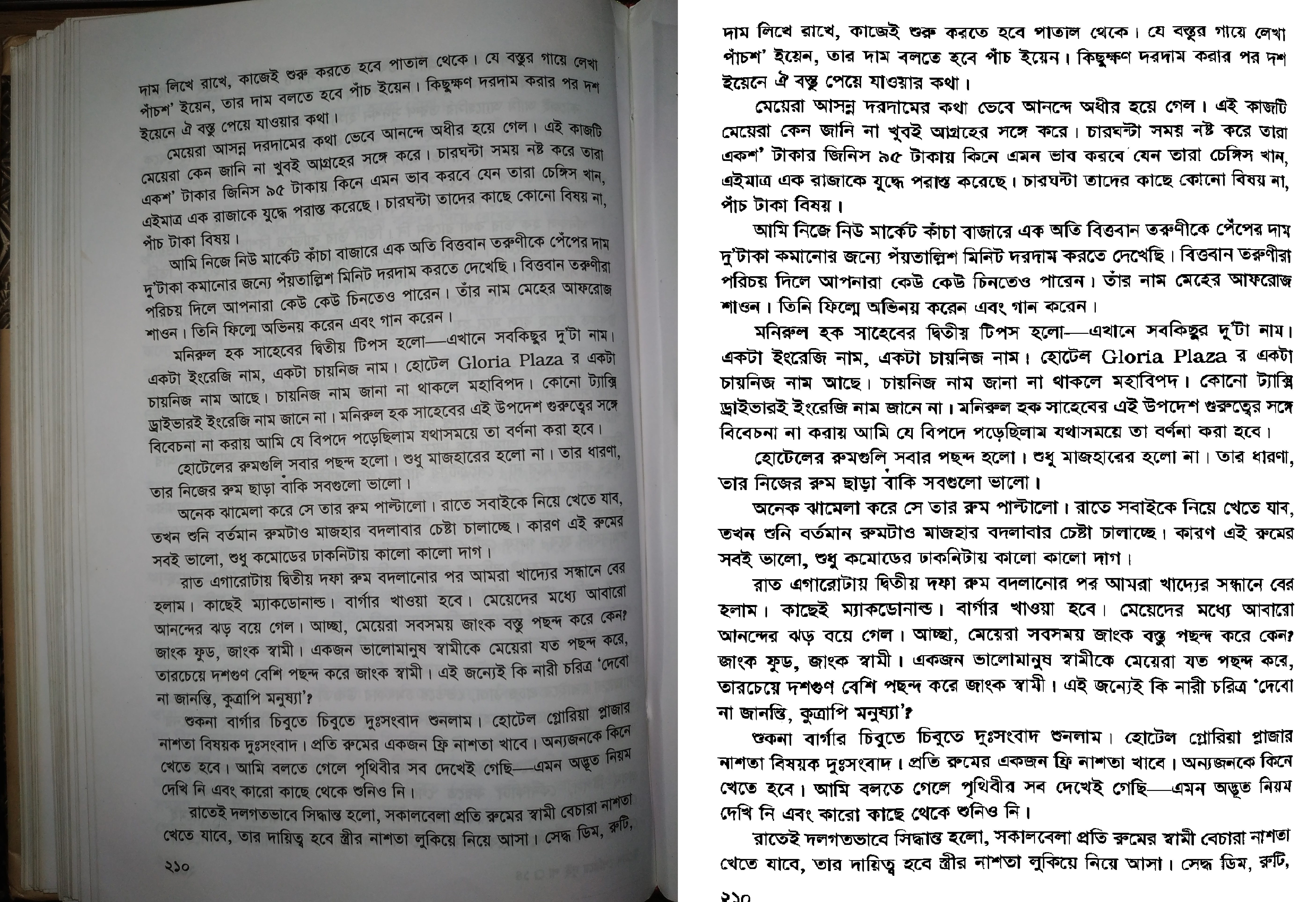}
  \caption{Example of sample image used (left) and the image being successfully pre-processed (right).}
    \label{fig: eg}
\end{figure}

Segmentation of Bangla text images provide results on segmented lines, words and characters. Our algorithm was run on 10 different Bangla text images of different fonts. The images were camera captured and contained constraints that our work has overcame to give good results. Fig. \ref{fig: eg} shows one example of our sample image alongside the image being successfully pre-processed by our methodology. The results are shown in Table \ref{tab: Result}. For line and word segmentation, we have been able to segment all the lines and almost all words accurately. For character segmentation, the result is convenient with 94.32\% accuracy. In spite of some limitations in our work, the accuracy level of line segmentation, word segmentation and character segmentation are remarkable. 

\begin{table}[ht]
  \centering
  \caption{Results of Line, Word and Character Segmentation.}
  \label{tab: Result}
  \begin{tabular}{lccc}
  \toprule
  Types of Segmentation&No. of Samples&No. Segmented&Accuracy (\%)\\
  \midrule
  Line Segmentation&312&312&100\% \\
  Word Segmentation&3525&3496&99.10\%  \\
  Character Segmentation&11208&10572&94.32\% \\ 
  \bottomrule
  \end{tabular}
\end{table}

Previous works on Bangla OCR system did not work with non-ideal cases. Their methods are not suitable for our sample images. To understand our contribution, we have left out subsets of our proposed non-ideal case handling steps and have performed experiments. The sample image in fig. \ref{fig: eg} was used to demonstrate this experiment. Table \ref{tab: Subset} shows effect in the accuracy of line, word and character segmentation when one of our proposed algorithm steps is excluded. 

The line, word and character accuracy of the multiple font step was not included from the image in fig. \ref{fig: eg}. The accuracy of this step was conducted from the image in fig. \ref{fig: MultipleFont}.
The accuracy comparison of each of the methods clearly shows the importance of each step and how all these steps together is the key for a better segmentation performance.

\begin{table}[ht]
  \centering
  \caption{Segmentation performance comparison has been shown both excluding different implemented steps and after inclusion of all steps.}
  \label{tab: Subset}
  \begin{tabular}{lccc}
  \toprule
  Excluded Step&Line&Word&Character\\
  \midrule
  Skew Correction&68.30\%&45.10\%&29.80\% \\
  Image Cropping&26.0\%&19.40\%&12.30\%  \\
  Image Dewarp&78.20\%&64.40\%&32.0\% \\
  Noise Reduction&84.0\%&67.40\%&47.0\% \\
  Multiple Font&92.0\%&87.20\%&76.40\% \\ \hline 
  With All Implemented steps&100\%&98\%&92.5\% \\
  \bottomrule
  \end{tabular}
\end{table}

We have not performed character recognition part in the current research. State-of-the-art Convolutional Neural Network (CNN) models are capable of performing multi-output classification task on input image if provided with properly labeled training samples (\cite{18}, \cite{19}, \cite{20}). We leave this data driven extension as part of future work. Our segmentation process does not always give perfectly segmented characters. Some common problems occur during removal of matra line and during character segmentation. 

\begin{figure}[ht]
  \centering
  \includegraphics[width=70mm]{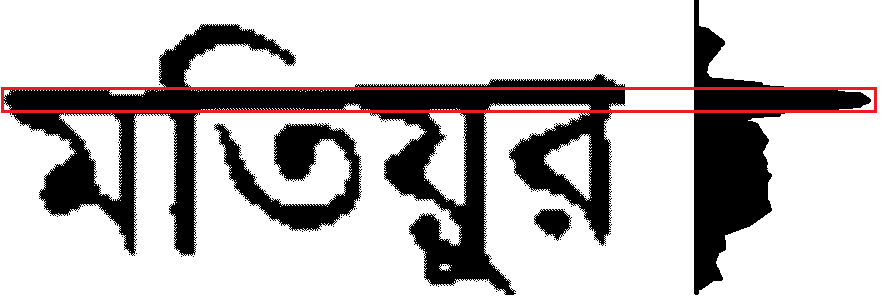}
  \caption{Region of Matra line Detected.}
    \label{fig: MatraHist}
\end{figure}

\begin{figure}[ht]
  \centering
  \includegraphics[width=70mm]{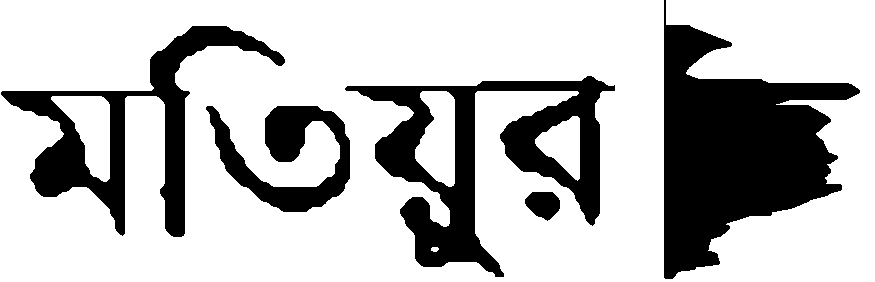}
  \caption{Region of Matra Line Undetected.}
    \label{fig: RemoveMatraHist}
\end{figure}

\begin{figure}[ht]
  \centering
  \includegraphics[width=\linewidth]{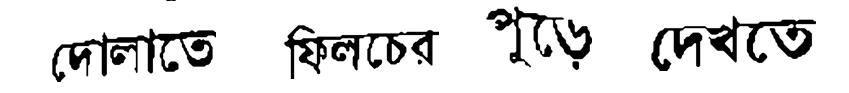}
  \caption{Matra Line Detection Failure Scenario Examples.}
    \label{fig: Limit2}
\end{figure}

When we straighten a curved line, a few words may stay tilted. In such cases, the matra line goes undetected, because the horizontal pixel sum criteria does not work here. This makes it harder for us to eliminate the matra line. Fig. \ref{fig: MatraHist} shows the region of matra line being detected of a slightly tilted word along with row-wise black pixel histogram. Fig. \ref{fig: RemoveMatraHist} shows the region of matra line which stays undetected and was not removed. As a result, we fail to segment such words into characters properly. Fig. \ref{fig: Limit2} shows some examples where the character segmentation of the words are not done correctly as removal of matra line was not successful.  

\begin{figure}[ht]
  \centering
  \includegraphics[width=\linewidth]{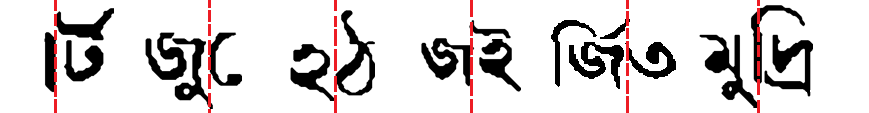}
  \caption{Character Overlap Example Scenario.}
    \label{fig: Limit3}
\end{figure}

Some of the Bangla printed characters overlap with each other depending on font type. It is difficult to segment such overlapping characters as there is no gap between them, which leads to poor character segmentation. If we try to segment such overlapping characters, some pixels from one character will be segmented with the other overlapping character. Fig. \ref{fig: Limit3} shows some of the failure cases in character segmentation due to overlapping of the characters.

\section{Conclusion}
This research aims at confronting the common challenges associated with non-ideal capture conditions of printed Bangla text image. As a pre-requisite of perfecting the segmentation phase of a Bangla OCR system, we have gone through a number of pre-processing steps. We have also identified and confronted common limitations of line, word and character segmentation phase. Properly segmented characters are the key to accurate recognition and reconstruction of the characters in digital media. Future researches may aim at resolving character overlap and matra line detection problem during character segmentation phase. For developing a fully functional Bangla OCR system, the current research has to be extended by developing a character recognition algorithm capable of identifying main character (vowel/ consonant/ compound character) and modified vowel from a segmented character image.  

\bibliographystyle{splncs04}

\bibliography{main}

\end{document}